\documentclass[sigplan,screen]{acmart}
\usepackage{multicol}
\usepackage{lipsum}
\renewcommand\footnotetextcopyrightpermission[1]{} 
\AtBeginDocument{%
  \providecommand\BibTeX{{%
    \normalfont B\kern-0.5em{\scshape i\kern-0.25em b}\kern-0.8em\TeX}}}

\setcopyright{acmcopyright}
\copyrightyear{2018}
\acmYear{2018}
\acmDOI{10.1145/1122445.1122456}

\acmConference[Woodstock '18]{Woodstock '18: ACM Symposium on Neural
  Gaze Detection}{June 03--05, 2018}{Woodstock, NY}
\acmBooktitle{Woodstock '18: ACM Symposium on Neural Gaze Detection,
  June 03--05, 2018, Woodstock, NY}
\acmPrice{15.00}
\acmISBN{978-1-4503-XXXX-X/18/06}

\makeatletter

\def\@copyrightspace{\relax}
\makeatother
\setcopyright{none}

\begin{document}

\title{
   GenderRobustness: Robustness of Gender Detection in Facial Recognition Systems with variation in Image Properties }
\subtitle {Preliminary Draft}

\author{Sharadha Srinivasan}
\affiliation{%
  \institution{Anna University}
  \streetaddress{1 Th{\o}rv{\"a}ld Circle}
  \city{Chennai}
  \country{India}}

\author{Madan Musuvathi}
\affiliation{%
  \institution{Microsoft Research}
  \city{Redmond}
  \country{USA}}



\begin{abstract}
In recent times, there have been increasing accusations on artificial intelligence systems and algorithms of computer vision of possessing implicit biases. Even though these conversations are more prevalent now and systems are improving by performing extensive testing and broadening their horizon, biases still do exist. One such class of systems where bias is said to exist is facial recognition systems, where bias has been observed on the basis of gender, ethnicity, skin tone and other facial attributes. This is even more disturbing, given the fact that these systems are used in practically every sector of the industries today. From as critical as criminal identification to as simple as getting your attendance registered, these systems have gained a huge market, especially in recent years. That in itself is a good enough reason for developers of these systems to ensure that the bias is kept to a bare minimum or ideally non-existent, to avoid major issues like favoring a particular gender, race or class of people or rather making a class of people susceptible to false accusations due to inability of these systems to correctly recognize those people.

In the last 2 years, facial recognition systems have claimed to have made their gender detection module more accurate and that they have trained the model on a more diverse dataset. Yet, in this paper, we present our findings on variations in gender detection of images when subjected to changes in image properties like brightness, sharpness, contrast as well as on a combination of these factors as opposed to just a single factor. We aim to test on different publicly available facial recognition systems for gender bias on a face dataset, which we have tried to make as diverse as possible in order to produce accurate results, and generate a comparative study.

\end{abstract}




\settopmatter{printacmref=false}
\maketitle
\pagestyle{plain}

\section{Introduction}
Facial recognition technology has come a long way since its inception about 50 years ago. Even though the first facial recognition system around the 1960s proved that it could be a viable biometric, it’s efficiency was greatly hampered by the technology of that era. It was as early as the 1970s that 21 specific subjective markers were introduced to automate the recognition. Especially in the past decade, it’s applications have become widespread. It’s uses go beyond just unlocking phones and laptops. Facial recognition systems now use artificial intelligence, machine learning and deep learning algorithms to predict a number of qualities from the detected faces. It’s applications have also been  growing exponentially due these features. It is widely used in law enforcement, retail, marketing and health applications, almost disrupting the way people live and work today, whether we realize it or not.

Can we concede with facial recognition software companies claiming that these models are able to accurately identify faces better than humans? On the contrary, to say that these systems are controversial would be an understatement. Unfortunately, it has been observed that facial recognition algorithms vary in their performance across different face types. As mentioned above, since law enforcement agencies have started using facial recognition to identify subjects, the accuracy of these algorithms is now a matter of criminal justice. Not just that, there have been several racial issues, starting with the 2009 face tracking software and prevalent in systems even today that can track fairer skin tones but not darker skin tones. This has led to several companies being called out, leading to outright bans of facial recognition software in specific places or for specific uses.

Recent studies and experiments showed that most of the popular facial recognition systems like Amazon Rekognition,Microsoft’s Cognitive Services Face API, Face++, Google Vision API and IBM’s Watson Visual Recognition API that identify a list of characteristics like gender, age, ethnicity, emotions and so on are gender biased. Following these accusations, Google and IBM have removed their gender recognition module from their systems. The main issue that was identified was that the training data used to train these models was not diverse enough. 

This paper is motivated by such biases, focussing on gender bias in facial recognitions systems. 
Though it has been established that these facial recognition systems are biased, there is a void in terms of exploration and analysis of gender bias patterns with variation in image properties. Would these facial recognition systems be as biased when the images are subjected to these transformations? If unbiased, then for any image the model should start with confidently correct, start showing ambiguity uniformly for all images or identify them correctly till the maximum value of the range.

Firstly, this paper explores the gender bias pattern when image transformations are applied on properties like sharpness, brightness and contrast. This paper also defines a robustness value to determine the range for which these experiments hold valid.

Secondly, Publicly available datasets like Adience, Labelled Faces in the Wild Home, Flickr-Faces-HQ Dataset (FFHQ) etc have proved to be inefficient while testing for bias due to their limited diversity. As we conduct experiments on image transformations, we also aim to create an all-inclusive dataset, to include every country on the globe. This would help us compartmentalize and study the bias of face recognitions systems with respect to different ethnicities.

\section{Implementation}
This tool, coded in python, is developed primarily for testing of the software for studying gender bias variations with variations in image
The model that we have chosen to start developing the tool for is the “Amazon Rekognition” software. 
\begin{itemize}
    \item To test on Amazon Rekognition, a Boto3 client is created that represents Amazon Rekognition and facilitates requests with the help of the AWS API. 
	\item A diverse dataset must be gathered in order to develop a testing tool that has appreciable efficacy. We start with a small set of images which included images from under-represented communities in the other publicly available datasets like South-East Asians, African Americans, South Asians and included images of other majority communities. 
	\item To test the Boto3 client, an image can be chosen at random from the dataset. The path is set and the image is read in “rb” mode which reads the image and it is then stored as a binary file. This byte file is sent as an argument to the “DetectFaces API” using the client to generate a response from the facial recognition software. 
	\item The response is essentially in the form of a dictionary, where the keys themselves are either a collection of dictionaries or a list of dictionaries. This response includes information about gender, age , emotion, moustache, smile, pose and various other details about the face, along with their confidence values.
\end{itemize}

This paper will focus on the accuracy and bias pertaining to the gender value and the corresponding confidence value.

\section{Image Transformations}

This paper explores the effect of applying image transformations on the gender value detected by the facial recognition system. The “ImageEnhance” function of the Pillow library used to perform these transformations. Pillow is a versatile and popular imaging library. The ImageEnhance module is a particularly useful module within the Pillow library. The ImageEnahance module consists of a number of classes that can be used for image enhancement. It contains in-built functions to change the brightness, sharpness, contrast and color of the image. While performing image transformations, the input and output, both would be in the form of images.

The image transformations discussed in this paper would be brightness, contrast and sharpness, as well as a combination of these factors. The gender bias becomes even more significant when the results are divided into the following three categories:
\begin{itemize}
    \item The gender value is correctly identified throughout all factors
    \item The gender value is incorrectly identified throughout all factors
    \item The gender value varies with variation in the factors
\end{itemize}
 \subsection{Brightness}
 The ImageEnahance module has a function called Brightness() where the brightness factor can be passed as a parameter to adjust the brightness of an image. Where ‘0’ would be a black and white image. When we call this function, what happens is the value of all pixels is changed by a constant. Adding a positive constant to all of the image pixel values makes the image brighter. Similarly, we can subtract a positive constant from all of the pixel values to make the image darker. 

We start with the factor ‘0’ all the way to ‘30’, with log intervals, to observe the changes in the gender values. The response is stored as a list of dictionaries. This is then used to visualize the results and find a range for the brightness factor and also observe any patterns in gender value variation. 

  \subsection{Contrast}
As mentioned above, brightness refers to the overall lightness or darkness of an image,whereas, contrast refers to the brightness difference between different objects or regions of the image. For this, the built in function of the ImageEnhance module Contrast() is used. To adjust the contrast, we can modify the slope of the transfer function. A transfer function with a slope greater than one will increase contrast, while a slope less than one will decrease it. 

We start with the factor ‘0’, which gives a solid grey image, all the way till 250, as some images are still able to be recognized at such high factors. The result thus obtained for each factor is stored and visualized to look for patterns and also find a range where the images are recognizable or show observable variations in gender values.
\subsection{Sharpness}
When the Sharpness() function is called, it increases the contrast between bright and dark regions to bring out features. During the sharpening process, a high pass filter is applied to an image.  An enhancement factor of 0.0 gives a blurred image, a factor of 1.0 gives the original image. Any factor above 1.0, increases the sharpness of the image.

Again, as in the other two factors, we start with the factor ‘0’, all the way till 200 after which the gender value becomes invariant or the image is not recognized by the facial recognition application.

\section{Dataset}
\section{Results}

\section{Conclusion}

\section{References}

\end{document}